\documentclass[sigconf, nonacm]{acmart}
\pdfoutput=1 

\renewcommand\footnotetextcopyrightpermission[1]{} 
\usepackage{xargs}                      
\usepackage[colorinlistoftodos,prependcaption,textsize=small]{todonotes}

\usepackage{enumerate}
\usepackage{url}
\usepackage{multirow}
\usepackage{graphicx}
\usepackage{pgfplots}
\pgfplotsset{width=8cm,compat=1.17}
\usepgfplotslibrary{colorbrewer}
\pgfplotsset{cycle list/Dark2-8}
\usepackage{enumitem}

\newcommand{\setword}[2]{%
  \phantomsection
  #1\def\@currentlabel{\unexpanded{#1}}\label{#2}%
}

\AtBeginDocument{%
  \providecommand\BibTeX{{%
    \normalfont B\kern-0.5em{\scshape i\kern-0.25em b}\kern-0.8em\TeX}}}

\begin{document}

\title{DeCorus: Hierarchical Multivariate Anomaly Detection at Cloud-Scale}

\author{Bruno Wassermann}
\email{brunow@il.ibm.com}
\affiliation{
  \institution{IBM Research Haifa}
  \city{Haifa}
  \country{Israel}
}
\author{David Ohana}
\email{david.ohana@ibm.com}
\affiliation{
  \institution{IBM Research Haifa}
  \city{Haifa}
  \country{Israel}
}
\author{Ronen Schaffer}
\email{ronen.schaffer@ibm.com}
\affiliation{
  \institution{IBM Research Haifa}
  \city{Haifa}
  \country{Israel}
}
\author{Robert Shahla}
\email{Robert.Shahla@ibm.com}
\affiliation{
  \institution{IBM Research Haifa}
  \city{Haifa}
  \country{Israel}
}
\author{Elliot K. Kolodner}
\email{kolodner@il.ibm.com}
\affiliation{
  \institution{IBM Research Haifa}
  \city{Haifa}
  \country{Israel}
}
\author{Eran Raichstein}
\email{eranra@il.ibm.com}
\affiliation{
  \institution{IBM Research Haifa}
  \city{Haifa}
  \country{Israel}
}
\author{Michal Malka}
\email{michal.malka@ibm.com}
\affiliation{
  \institution{IBM Research Haifa}
  \city{Haifa}
  \country{Israel}
}

\renewcommand{\shortauthors}{Wassermann and Ohana, et al.}

\begin{abstract}
  Multivariate anomaly detection can be used to identify outages within large
  volumes of telemetry data for computing systems. However, developing an
  efficient anomaly detector that can provide users with relevant information is
  a challenging problem.  We introduce our approach to hierarchical multivariate
  anomaly detection called DeCorus, a statistical multivariate anomaly detector
  which achieves linear complexity. It extends standard statistical techniques
  to improve their ability to find relevant anomalies within noisy signals and
  makes use of types of domain knowledge that system operators commonly possess
  to compute system-level anomaly scores.  We describe the implementation of
  DeCorus an online log anomaly detection tool for network device syslog
  messages deployed at a cloud service provider.  We use real-world data sets
  that consist of $1.5$ billion network device syslog messages and hundreds of
  incident tickets to characterize the performance of DeCorus and compare its
  ability to detect incidents with five alternative anomaly detectors.  While
  DeCorus outperforms the other anomaly detectors, all of them are challenged by
  our data set. We share how DeCorus provides value in the field and how we plan
  to improve its incident detection accuracy.  
\end{abstract}





\maketitle

\section{Introduction}
\label{introduction}


The gold standard of \textit{high availability} for cloud service providers are
the \textit{five nines} of availability: an uptime of at least $99.999\%$. This
translates to a maximum permissible downtime, scheduled or otherwise, of less
than $30$ seconds per month. This is an exceedingly difficult goal for any
large, complex computing system and failing to achieve it may result in negative
business impact in the form of reduced customer satisfaction, loss of reputation
and, ultimately, a loss of revenue, not only for the cloud service provider, but
also for the businesses which build their applications on top of its services.

Monitoring is a crucial ingredient in achieving high availability. We are able
to collect vast amounts of data about the operation of a computing system. A
variety of tools (e.g.,~\cite{elasticsearch},~\cite{prometheus},~\cite{jaeger})
allow for the collection of log events, metrics and request traces. The data
collected by these tools are essentially time series that capture different
aspects of the behavior of the processes executing in our computing systems. The
general expectation is that service disruptions can become visible in such
monitoring data as data points that are significantly different from a majority
of other data points, so-called \emph{anomalies}. 

Failure detection approaches that require a significant amount of manual effort
are not suitable at scale. The infrastructure of a single cloud data center
consists of thousands of components, for each of which we may measure large
numbers of metrics and log events. Summarizing this data in the form of
dashboards, trying to define static thresholds over monitoring metrics or rule-
and model-based approaches~\cite{244794,Kliger1995} do not provide effective
ways to detect failures accurately and quickly.  An instance of the latter
approach we have encountered in practice is a large set of rules that specify
what network device syslog message should raise an alert.  These rules are
maintained by a team of Network Reliability Engineers (NREs) for whom we build
tools. The deterministic rules detect known failures with high accuracy, but
require ongoing maintenance and tend to detect known failure types.  There is
scope for a more automated way to detecting failures or incidents\footnote{The
NREs use the term incidents as some issues only lead to a reduction in
redundancy that needs to be dealt with, but does not result in user-visible
outages.}.

Finding anomalies in monitoring data is one way to automate the detection of
known and unknown failures.  There are many approaches to univariate anomaly
detection (UVAD)\cite{Chandola2009}. These methods can detect anomalies in an
individual metric, or in our context, in a time series that represents the
measurements of a single metric over time.  Basing alerts on univariate
anomalies may be counter-productive as problems often affect more than just a
small set of monitoring metrics. Raising an alert for each univariate anomaly,
would overwhelm NREs. For anomaly detection to be useful as a failure alerting
mechanism in large systems, it needs to be able to detect anomalies in multiple
time series simultaneously.  Existing multivariate anomaly detection (MVAD)
approaches typically fall into one of several categories, which we review in
Section~\ref{motivation:mvads}. We find that many techniques have shortcomings
when evaluated against some of the requirements for incident detection in large
systems that we have learned about across several use cases and briefly describe
in Section~\ref{motivation:requirements}. 

We have developed~\textit{DeCorus} as a statistics-based MVAD that operates on
temporal data. It addresses many of the identified requirements.  First, it is
\emph{computationally efficient} in that it achieves linear runtime and space
complexity in the number of data points.  This makes DeCorus a
\emph{cost-efficient} option for detecting incidents in the infrastructure of a
large cloud computing provider.  Second, DeCorus is able to learn from
monitoring data without requiring manual curation of nominal reference data or
labeled data (\emph{unsupervised}). Third, DeCorus automatically adapts its
anomaly detection model to changes in the system, which it does by virtue of
using statistical techniques with a memory component that assigns increasing
weights to more recent measurements (\emph{online learning}).  It is able to
make use of temporal characteristics of the data (\emph{temporal-aware}). And
finally, it is able to correlate a top-level anomaly score with the
contributions of individual anomalies at lower layers of the system
(\emph{hierarchical aggregation}), which enables it to provide hints about root
causes. 

The hierarchical aggregation DeCorus performs is based on two types of domain
knowledge that are readily available. The first type of domain knowledge is a
high-level view of system structure. In practice, operators often have a basic
model of the hierarchical composition of a system.  The second type of domain
knowledge is the relative criticality of components with regard to their impact
on service reliability. DeCorus makes use of this information to aggregate
anomalies from individual components into a system-level anomaly score refined
by explicit weights. 

In this paper, we describe the implementation of DeCorus as an online log
anomaly detection tool for network device syslog messages in the data centers of
a cloud service provider.  DeCorus serves as a complementary monitoring tool to
an existing array of solutions and assists the NREs by directing their attention
to interesting issues.

The contributions of this paper are as follows:
\begin{itemize}
  \item{List of the requirements of large-scale uses cases for
  MVAD communicated by end users.}
  \item{An exposition of our approach to hierarchical anomaly detection.} 
  \item{An experimental evaluation based on a production data set of
  network device syslog messages and incidents from a large cloud
  computing provider, consisting of:}
  \begin{itemize}
    \item{characterization of the resource utilization and runtime of the core algorithms in DeCorus;}
    \item{a comparative evaluation of the incident detection accuracy of DeCorus and five alternative anomaly detection techniques.}
  \end{itemize}
\end{itemize}

In Section~\ref{motivation:requirements}, we discuss requirements for anomaly
detection for effective identification of incidents in large systems. We review
existing anomaly detection approaches in Section~\ref{motivation:mvads}.  In
Section~\ref{decorus}, we provide an overview of the implementation of DeCorus
in the form of a log anomaly detection pipeline, before discussing the techniques
used in its core algorithms.  We present the experimental evaluation of DeCorus
in Section~\ref{experiments}. Finally, after reviewing closely related work, we
discuss our conclusions and plans for future work.

\section{Cloud-scale MVAD}
\label{motivation}

The requirements we briefly share in this section are based on insights gained
from working with teams of Site Reliability and Network Reliability Engineers
(SREs/NREs) across a number of production use cases.

\subsection{Requirements}
\label{motivation:requirements}

The ability to detect failures close in time to their occurrence requires
\textbf{low processing latency} (\setword{Req. I}{req:processinglatency}).  It
is important to enable SREs/NREs to repair failures quickly. 

The second requirement is \textbf{cost-efficient processing}
(\setword{Req. II}{req:cost}).  The cost of running the algorithms needs to be
substantially lower than the cost savings from any downtime they reduce. 

In order to achieve low failure detection latencies and cost-efficient
application in large systems,~\textbf{computational efficiency} (\setword{Req.
III}{req:efficiency}) is needed to handle large volumes of monitoring data.

The fourth requirement is for \textbf{temporal-aware data handling}
(\setword{Req. IV}{req:temporal}).  Monitoring data is represented as time
series. Data points should be evaluated within their temporal context and
sustained anomalies should be given more weight.

The fact that data is generated as a boundless stream gives rise to the fifth
requirement of techniques being able to process data using \textbf{stream
processing} (\setword{Req. V}{req:stream}). 

When things go wrong in a large system, it is often the case that many processes
in multiple components are affected. An anomaly detector needs to
\textbf{identify system-level anomalies} (\setword{Req. VI}{req:systemlevel})
that arise from multiple individual time series. 

In spite of the need to alert on system-level anomalies, we do not want to lose
the ability to \textbf{identify the contributions of individual anomalies to
top-level anomalies} (\setword{Req.  VII}{req:anomalycontributions}). 

NREs need to be able to trust that many incidents are detected reliably. This
gives rise to the eighth requirement for \textbf{highly accurate alerting}
(\setword{Req.  VIII}{req:accuratealerts}). 

A requirement for \textbf{unsupervised learning} (\setword{Req.
IX}{req:unsupervised}) may be in conflict with the requirement of high accuracy.
However, curation of training data sets can be labor-intensive and we should
strive to limit it.

The ability for \textbf{online learning} (\setword{Req. X}{req:online}) allows
an anomaly detector to adapt to changes automatically without mechanisms to
detect drift and perform retraining and validation.

Some components in a large system have more impact on availability than others.
A failure detection mechanism should be able to make use of such knowledge in
order to \textbf{weight anomalies by component criticality} (\setword{Req.
XI}{req:criticalityweighting}). 

We may collect more metrics for some components than we do
for others. It would be beneficial to be able to \textbf{normalize the
contributions of components to anomalies} (\setword{Req.
XII}{req:implicitweights}) so that the number of metrics available for a
component does not by itself influence its contribution to the system-level
anomaly score. 

We do not expect any single technique to be able to meet each one of these
requirements, but this list can represent a set of goalposts as they have been
provided by users responsible to maintain several large production systems

\subsection{Existing MVAD approaches}
\label{motivation:mvads}

We keep this review concise and refer the interested reader to the provided
references and to~\cite{chalapathy2019deep} for anomaly detection based on Deep
Learning.

\textbf{Neighbor-based approaches.} The main idea behind nearest-neighbor based
techniques for anomaly detection is that most data falls within regions of high
density forming so-called neighborhoods of data points that are close to one
another.  A representative technique of the density-based approach is Local
Outlier Factor (LOF)~\cite{10.1145/335191.335388}. LOF computes the density of a
data point by finding the radius of a hypersphere around it that
includes its $k$ nearest neighbors.  A data point whose density is significantly
less than that of its neighbors, is considered to be an anomaly.

Nearest-neighbor based techniques work in an unsupervised manner, can handle
multivariate data and suitable distance functions for temporal data
exist~\cite{bellman1959adaptive}. It should be possible to weight the
computation of the anomaly score using available domain knowledge. While
neighbor-based techniques generally exhibit quadratic runtime in the number of
data points, there are techniques that improve their efficiency by approximation 
~\cite{10.1145/502512.502554} or prefix filtering~\cite{10.1145/1242572.1242591}.

\textbf{Clustering-based approaches.} Clustering-based anomaly detection groups
data points into clusters of similar instances based on some notion of distance.
The anomaly score of a data point can be based on the distance to its closest
cluster.  Microsoft's LogCluster~\cite{lin:mslogcluster:2016} transforms raw
logs into unique events that are grouped into log sequences based on a
transaction identifier present in the logs.  The log sequences are grouped into
clusters using Agglomerative Hierarchical
clustering~\cite{https://doi.org/10.2307/2346439} and the most representative
log per cluster is identified.   This approach can be adapted for anomaly
detection by flagging those logs whose distance from their cluster's
representative log exceeds a threshold. 

Clustering algorithms work in an unsupervised manner and can be applied to
multivariate data with the use of suitable distance functions.  However, there
are shortcomings.  First, Agglomerative Hierarchical
clustering~\cite{https://doi.org/10.2307/2346439} typically incurs $O(n^3)$
runtime complexity and quadratic space overhead in the number of data points,
but linear-complexity algorithms (e.g.,~\cite{pakhira:kmeans:2015}) exist.
Second, clustering algorithms cannot take temporal relationships between data
points into account. 

\textbf{Classification-based approaches.} We focus on unsupervised techniques.
One-Class Support Vector Machines (OC-SVMs)~\cite{shoelkopf:ocsvm:2001} learn
decision boundaries that surround high-density data and allow the classifier to
distinguish data in the target class from data points outside of it.  OC-SVMs
can be used in an unsupervised manner.  The second technique is Isolation
Forests~\cite{liu:isolationforest:2008}, which do not generate a profile of
normal data to detect deviations from that profile, but instead exploit the
insight that anomalous data points occur in much smaller proportion than normal
data points. Isolation Forests split the data recursively to identify those
points that require fewer splits to isolate and can be considered outliers.

When used in an unsupervised manner, the accuracy of OC-SVMs depends on a
parameter that captures the expected proportion of outliers that will be present
in the data. This can be difficult to estimate. While the complexity of SVMs
depend on several factors, typically their runtime complexity is cubic in the
number of data samples. Isolation Forests have linear runtime complexity, which
makes them well-suited to large data sets. However, they are only able to detect
point anomalies.

\textbf{Subspace approaches.} Subspace-based techniques 
project a highly-dimensional data set into a lower-dimensional subspace, in
which it may be easier to distinguish normal data points from anomalous ones.
In~\cite{shyu:novel:2003}, the authors apply anomaly detection based on
Principal Component Analysis (PCA)~\cite{Jolliffe:pca:2002} for network
intrusion detection. The principal components are computed on anomaly-free data
and components that capture $50\%$ of variation are selected as the major ones.
New data points whose distance from major components is greater than a threshold
are considered anomalous. In ~\cite{xu:2009}, the authors describe how they
parse application console logs into event count vectors and then apply PCA to
detect anomalies.

Subspace-based approaches can work in an
unsupervised manner (e.g., using robust PCA~\cite{huber:robust:2004}), and
work with multivariate data by definition. However, there are
issues when handling large volumes of streaming data.  Principal
components are computed once and then relied upon to evaluate incoming data
points. To handle temporal data, the principal components might need to
be recomputed periodically. Furthermore, PCA requires the covariance matrix for
all variables to be kept in memory. There has been work on implementing PCA in
an online manner~\cite{lee:onlinepca:2013}.  Finally, the computational
complexity of PCA is typically linear in the number of data points and cubic in
the number of attributes~\cite{Chandola2009} ($O(p^2n + p^3)$, where $n$ is the
number of data points and $p$ is the number of attributes per data point). 

\textbf{Statistical approaches.} Statistical anomaly detection works by fitting
a statistical model to existing data and performing a test on incoming data
points to decide whether they belong to the observed model.  There are
parametric techniques that make assumptions about the underlying distribution of
the data and non-parametric ones.  An example of a parametric technique is
described in~\cite{tsay2000outliers} which extends statistical tests for
regression models to multivariate data. An example of a non-parametric technique
is HBOS~\cite{Goldstein2012} which builds a histogram based on existing data
points and checks if new data points fall within one of the bins.  A more
comprehensive review can be found in~\cite{Chandola2009}.

Statistical anomaly detection approaches offer several benefits. There are 
models that can be fit to the data with linear complexity.  They typically work
in an unsupervised manner. The output of statistical inference tests can be used
to provide meaningful anomaly scores and the scores can be scaled to account for
domain knowledge.  There are also a few disadvantages.  Parametric techniques
assume a specific distribution.  Histogram-based techniques are not able to take
relationships between attributes into account.  

Statistical techniques seem to provide a good basis for multivariate
anomaly detection on large data sets. The question is whether the shortcomings
of the other approaches reviewed are justified by increased failure detection
accuracy compared with statistical approaches when applied to real-world data.

\section{DeCorus}
\label{decorus}

\subsection{Log Anomaly Detection Pipeline}
\label{decorus:logad}

\begin{figure*}
  \includegraphics[width=1\textwidth]{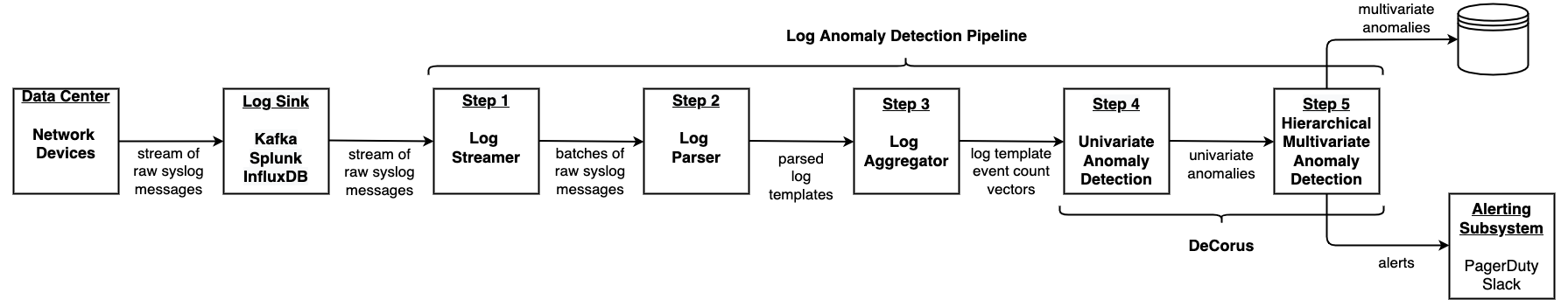}
  \caption{The stream of syslog message generated by the network devices are processed by the five steps of our log anomaly detection pipeline. Steps 4 and 5 implement our algorithms for univariate and hierarchical multivariate anomaly detection.}
  \label{fig:decorus:logadpipeline}
\end{figure*}

We have implemented DeCorus as part of a streaming log anomaly detection
pipeline. This pipeline ingests syslog messages that are generated by the
network devices comprising the infrastructure of a data center. It
converts the raw log messages into time series that represent counts of how
often different types of events have occurred and then applies the DeCorus
anomaly detection algorithms to this data to compute an anomaly score for the
overall system, a data center, and its sub-components, such as redundancy groups
of devices, individual network devices and their log events.

The diagram in Figure~\ref{fig:decorus:logadpipeline} gives an overview of this
pipeline.  The network devices are routers and switches that produce a stream of
syslog messages. The syslog messages are submitted into a so-called~\textbf{Log
Sink}, which persists the incoming log messages and makes them available to
consumers.  Examples of sinks our log anomaly detection pipeline has been
integrated with are Splunk~\cite{splunk}, InfluxDB~\cite{influxdb} and Apache
Kafka~\cite{apache:kafka}. The~\textbf{Log Streamer} component (\textbf{Step 1}
of the pipeline) retrieves data from the sinks and prepares raw log messages for
subsequent steps. The~\textbf{second step} in the pipeline is the~\textbf{Log
Parser} which encapsulates logic to parse different log formats and uses an
implementation~\cite{ibm:drain3} of the DRAIN algorithm~\cite{he:drain:2017} in
order to identify the unique types of log messages. These so-called~\emph{log
templates} correspond to different types of events.  In~\textbf{Step 3},
the~\textbf{Log Aggregator} computes a count of how often each type of log
message has occurred recently. It creates one time series per unique combination
of network device and log template. We also refer to these time series
as~\emph{event count vectors}.  The event count vectors serve as input to
the~\textbf{Univariate Anomaly Detection} algorithm (\textbf{Step 4}). The
anomalies it identifies in the individual time series are the input to
our~\textbf{Hierarchical Multivariate Anomaly Detection} algorithm (\textbf{Step
5}), which computes the overall system and sub-component anomaly scores on an
ongoing basis and raises alerts.

\subsection{Univariate Anomaly Detection}
\label{deorus:logad:uvad}

The univariate anomaly detector (UVAD) receives one time series per log template
(event type) on a network device as input and produces an anomaly score for the
latest data points as output. It uses an improved Z-Score to compute 
anomaly scores. Z-Scores measure the number of standard deviations that a data
point is away from the rest of the observed data. A common approach is to
interpret any data point with a Z-Score of $3.0$ or greater as an anomaly.
DeCorus applies several improvements to handle noisy signals. First, it
compares the anomaly score in a small recent window (e.g., 5 minutes)
to the anomaly score for the same signal in a larger overlapping window (e.g.,
24 hours), both computed using an exponentially weighted moving average (EWMA).
An anomaly score that is similar to anomaly scores observed over the past 24
hours is less informative than one whose current anomaly is significantly higher
than what has been observed in the larger window.  This provides some additional
context and is referred to as Gaussian Tail
(\cite{DBLP:journals/corr/AhmadP16,DBLP:journals/corr/abs-1708-03665}).  Second,
it attempts to give more weight to continuous anomalies to avoid flagging
point anomalies.  It boosts the anomaly scores of a signal in proportion to the
number of continuous anomalous data points.  These are two improvements that
DeCorus makes to standard Z-Score computations.

A new type of event occurring on a network device that has been monitored for
several months could be considered an interesting anomaly. Our UVAD boosts the
anomaly score of previously unseen log templates by treating the data points of
the signal at times prior to detection of this new template to have values of
zero. Any non-zero count after a long period of zeroes will lead to an increased
anomaly score for this new log template, until its non-zero values become more
prevalent.  Another feature of our UVAD is its ability to use metadata about
metrics. It accepts value bounds per signal and can be configured to detect only
positive or negative anomalies or both.

The runtime complexity of UVAD is linear in the number of data points. UVAD
performs a one-time setup for each new time series, which mainly consists of
zero-padding signals for time stamps that occurred prior to the occurrence of a
new log template. These operations are simple.  The bulk of the computational
overhead is incurred by the Z-Score computations, which have linear complexity
in the number of data points. 

\subsection{Hierarchical Multivariate Anomaly Detection}
\label{decorus:logad:mvad}

The hierarchical multivariate anomaly detection (HMVAD) algorithm in DeCorus
computes a single multivariate anomaly score for a time window. It makes use of
two types of domain knowledge. One is knowledge about system structure
which is used to build a topological tree that describes the monitored system as
a hierarchical composition of sub-components and their event types.  The other
is a set of weights that provide an indication of their importance for the
reliable operation of the system.  HMVAD computes an anomaly score for each node
in this tree and the anomaly score at the root of the tree is considered to be a
measure of how anomalous the overall system (e.g., an entire data center) has
been recently.  

HMVAD performs the following main steps:
\begin{enumerate}[topsep=1ex]
  \item Collect the anomalous data points of all time series in the
  current time window that were identified by UVAD.
  \item Filter out anomalous data points according to signal-specific
  configuration (e.g., devices matching a prefix).
  \item For each time series (log template on a network device), select 
  the anomalous data point in the current time window with the maximum anomaly 
  score.
  \item Update the leaf nodes of the topological tree with the anomaly score
  of the corresponding data points from the previous step.
  \item Compute the anomaly score of the root node and all child nodes by
  post-order traversal (bottom-up) of the topological tree. 
  \item Calculate the percentile rank for each node based on its anomaly
  score and the historic scores of all nodes at the same level in the
  topological tree.
  \item Compute a human-friendly anomaly score by normalizing the percentile
  ranks above a threshold into the range $[0, 100]$.
\end{enumerate}
HMVAD uses two types of input. 
The anomalous data points that UVAD has identified in the individual
time series are provided as a triple consisting of 
\verb|(timestamp, anomaly score, dimensions)|, where~\emph{dimensions} are 
key-value pairs that uniquely identify the source of the signal. For log anomaly
detection in networks, the
dimensions consist of a network device name and the identifier of the log
template.

The second type of input is metadata about the system.  NREs often have a basic
model of the hierarchical composition of a system.  This model does not
necessarily lend itself to create a complete dependency graph, but results in a
topological tree whose root node represents the overall system being monitored.
The leaf nodes correspond to individual log templates (event types).  The
intermediate nodes represent components or groupings of components that are
defined by the NREs.  In the example depicted in
Figure~\ref{fig:decorus:networkhierarchy}, the root node represents an entire
data center network. Beneath the data center are redundancy groups of certain
types of network devices as well as individual network devices. Devices are
grouped by their role in the system (e.g., \verb|role-1| could be a frontside
customer router).  Finally, the leaf nodes represent the log templates generated
by their parent nodes.  Domain experts can choose any structure that can be
represented as a tree. 

There tends to be consensus on which components are the most and the least
important ones. We have found that this extends to
knowledge about which log templates are more indicative of a relevant issue. In
DeCorus, this knowledge is encoded in the form of real-numbered weights in the
range $[0.0, 100]$.  Components and event types with higher weights have their
anomaly scores boosted accordingly. 

\begin{center}
  \includegraphics[height=45mm]{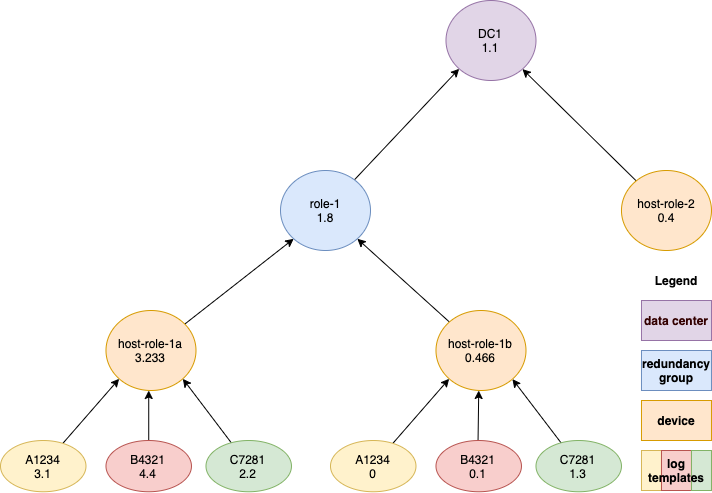}
  \captionof{figure}{Part of a topological tree for a data center network
  with network device redundancy groups, individual devices and log
  template identifiers.
  The values are anomaly scores.}
  \label{fig:decorus:networkhierarchy}
\end{center}

HMVAD correlates anomaly scores based on the time of the anomaly and the source
component of the signal. \emph{Temporal correlation} is performed by considering
all anomaly scores within a sliding time window of size $d$ and step size $s$
and selecting the maximum anomaly score for each unique signal as the
representative data point in this time window. The use of a sliding window
ensures that anomalies close to the edges of intervals are not missed during
correlation.  Figure~\ref{fig:decorus:aggregationtemporal} illustrates the
temporal aggregation. The window size is $15$ minutes with an overlapping time
window beginning every $5$ minutes. There are three unique signals in this
example denoted by the colored circles. The numeric values are the univariate
anomaly scores computed by UVAD.  HMVAD aggregates the anomaly scores multiple
times per time window, which allows it to detect significant anomalies before
the end of an interval.

\begin{center}
  \includegraphics[width=80mm]{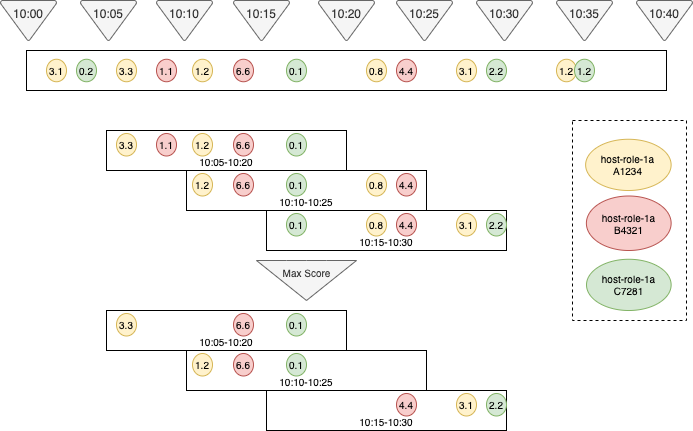}
  \captionof{figure}{HMVAD collects the anomaly 
  scores for each signal in 15-minute windows with new overlapping windows
  started every $5$ minutes and selects the maximum anomaly score per signal
  in each time window.}
  \label{fig:decorus:aggregationtemporal}
\end{center}

\emph{Spatial correlation} of anomaly scores is performed with the help of the
topological tree representation of the system.  During each time window, the
anomaly scores are aggregated in a post-order traversal starting at the leaf
nodes. The anomaly score at a parent node (e.g., network device) is the result
of the aggregation of the anomaly scores of its child nodes (e.g., events).  

HMVAD uses a Weighted Power Mean (WPM)~\cite{bullen:weightedmeans:2014} as the
aggregation function, which enables HMVAD to incorporate weights that NREs
provide for components and log templates.  The actual anomaly score for a signal
is a combination of all relevant weights using a squared geometric mean that is
normalized to the range $[0, 1]$.  While each weight is normalized to the range
$[0, 1]$, we do not normalize the sum of the weights. We do so to enable
comparisons between the anomaly scores of sibling nodes in the tree.  For
example, let us have two nodes~\emph{A} and~\emph{B} that both have four child
nodes. The child nodes of node~\emph{A} each have a weight of $0.2$ and the
child nodes of node~\emph{B} each have a weight of $0.4$. Assuming equal raw
anomaly scores, using normalization of the sum of the weights both~\emph{A}
and~\emph{B} will have the same aggregated anomaly score, even though the child
nodes of node~\emph{B} are considered to be twice as important.  Our use of WPM
allows HMVAD to ensure that the aggregated anomaly score reflects the domain
knowledge.

In the absence of explicit weights, HMVAD aggregates anomaly scores using the
topological tree representation to compute weights implicitly.  It implicitly
assigns an equal weight for every immediate sub-node of the same parent node, no
matter how many children exist below that sub-node. This helps to overcome 
situations where some sub-components produce more signals than others and avoids
such components unduly influencing the system-level anomaly score.

The final step in the aggregation is to normalize the anomaly score at each node
into the range $[0.0\%, 100\%]$.  This normalization is performed by computing
the percentile rank of the latest anomaly score in comparison to all historic
values of a reference group, which consists of all nodes at the same depth of
the topological tree. Even a relatively small change to the anomaly score of a
network device that has acted normally so far, could gain a high percentile
rank. However, if we examine its anomaly score within the context of all other
network devices at the same level in the tree, we may find that its behavior is
within typical bounds for devices of its type. Any percentile rank value below a
given threshold is transformed to $0$ and scores above the threshold are
linearly transformed to fill the range from $0$ to $100$. Using percentile ranks
in this way focuses the results on the most significant anomalies.

\subsection{Suitability of DeCorus}
\label{motivation:novelty}

DeCorus satisfies most of the requirements that we have identified in
Section~\ref{motivation:requirements}.  First, ANON enables large-scale log
anomaly detection in a cost-efficient manner (\ref{req:cost}) thanks to its core
processing algorithms achieving linear complexity in the number of data points
(\ref{req:efficiency}). In a production deployment, each step of our log anomaly
detection pipeline consumes well below 1 GB of RAM on average and little CPU.
Second, the algorithms in DeCorus make use of some of the temporal
characteristics of the data (\ref{req:temporal}) by comparing anomalies in the
current time window with the anomalies in a larger background window and by
boosting the anomaly scores of signals with continuous anomalies.  Third,
DeCorus is able to process the data in an online manner (\ref{req:stream}). Our
HMVAD algorithm is able to compute a system-level anomaly score for multivariate
anomaly detection (\ref{req:systemlevel}) by making use of a topological tree
representation of the system and the use of suitable aggregation functions,
while still maintaining the ability to identify the contributions of
sub-components and event types (\ref{req:anomalycontributions}).  Fourth, the
core anomaly detection algorithms in DeCorus are based on statistical techniques
with a configurable memory component that assigns increasing weights to more
recent measurements, which allows DeCorus to adapt to changes in the underlying
system (\ref{req:online}).  Finally, DeCorus is able to take knowledge about the
importance of different types of components into account 
(\ref{req:criticalityweighting}) and even in the absence of explicit weights, it
normalizes (\ref{req:implicitweights}) the contributions of different components
so that no single component can influence the anomaly score merely based on the
number of event types originating at it.

There are requirements that DeCorus achieves partially. First, it raises an
alert for a failure within five minutes of its occurrence, but it is not able to
detect failures accurately within seconds (\ref{req:processinglatency}).
Second, while its algorithms are unsupervised, DeCorus can benefit from having
some of its configuration parameters (e.g., memory parameters for moving
averages, durations of background time windows) tuned for the system under
monitoring. Finally, ANON could make better use of temporal relationships in the
data (\ref{req:temporal}).

\section{Experimental Evaluation}
\label{experiments}

The main research questions we address are:

\begin{itemize}
  \item{\textbf{RQ 1: Performance.} In terms of its computational overhead, can
  DeCorus be applied to large systems that produce a high volume of log messages?}
  \item{\textbf{RQ 2: Accuracy.} What is the failure detection accuracy DeCorus
  achieves on~\emph{real-world data sets} and how does it compare to 
  alternative approaches?}
\end{itemize}

\subsection{Performance}
\label{experiments:performance}

To examine the claim that DeCorus achieves linear scalability, we quantify the
computational requirements of its core algorithms for univariate and
multivariate anomaly detection (\emph{UVAD} and~\emph{MVAD}). We measure their
resource utilization in terms of~\emph{memory consumption} in MB and~\emph{CPU
utilization} in percent, where $100\%$ represents full utilization of a single
core. We measure the runtime of the algorithms as the~\emph{wall clock time} in
seconds spent on core processing activities and~\emph{throughput} as time spent
per $1,000$ data points. Please note that while DeCorus has been implemented as a
stream processing system, the other approaches we evaluate are batch processing
systems and as such their performance measurements would not be comparable to
that of DeCorus. 

We want to observe the core algorithms.  To reduce the time spent waiting for
the previous processing steps and on I/O, we pre-process all available log
messages by running them through the pipeline steps for log parsing, log
template extraction and conversion into time series. We write the resulting
event count vectors, into an Apache Kafka instance in their entirety. Second, we
execute the UVAD step of the pipeline repeatedly (10 times) on the event count
vectors and then run the MVAD step repeatedly on its output.  This increases
resource utilization as much as possible.

The experimental setup consisted of a VM with an Intel Xeon 8260 CPU clocked at
$2.40$GHz with eight cores, $64$GB physical memory and a $2$TB of
network-attached storage backed by SSDs.  The UVAD and MVAD algorithms were
executed in their respective Docker containers.  We measured the resource
utilization metrics using the Docker stats API~\cite{dockerstatsapi}, and
instrumented the code to measure the wall clock time of key steps in the
algorithms. 

The two data sets for the performance experiments consist of the syslog messages
emitted by the routers and switches in a data center of a cloud service
provider.  We have collected $149$ million log messages generated by $402$
devices over a period of $200$ days. Our pre-processing turned these log
messages into more than $7,000$ time series signals comprised of more than $280$
million data points.  In order to examine the scalability behavior of DeCorus,
we have doubled the number of data points per unit of time by adding dummy host
and log template identifiers for every existing entry.  We refer to the
resulting two data sets as~\emph{DC-Z-100\%} and~\emph{DC-Z-200\%} respectively.

\subsubsection{Resource Utilization}
\label{experiments:performance:resources}

Table~\ref{table:performance:memory} compares the mean and peak memory
consumption of the core algorithms in DeCorus.
Table~\ref{table:performance:cpu} presents the measurements of CPU utilization.
Given that the algorithms perform mainly sequential processing, we do not expect
their resource utilization to increase significantly when the number of data
points double.  The tables list the measurements for both data sets for
completeness.

First, we review the memory and CPU consumption of the UVAD algorithm. We find
a mean memory consumption of $446.93$MB (95\% CI: $[444.53,
449.33]$) for DC-Z-100 and only a relative increase of about $6.81\%$ for the
data set with twice as many data points. The observed peak memory utilization
was $604.22$ MB for DC-Z-100 and an increase of about $11\%$ for the larger data
set. Mean CPU utilization of UVAD was $148.40\%$ (95\% CI: $[147.96, 148.84]$),
which corresponds to about $1.5$ CPU cores being used on average. The peak CPU
utilization was $614.27\%$ for the smaller data set with a $7.7\%$ relative
increase for DC-Z-200.

Next, we examine the resource utilization of the MVAD algorithm. We observe a
mean memory consumption of $207.76$ MB (95\% CI: $[205.00, 210.51]$) that
increases by about $15\%$ for the larger data set and peak memory consumption of
$418.22$ MB. The mean CPU utilization was $98.23\%$ (95\% CI: $[95.34,
101.12]\%$) and peak CPU utilization was $322.57\%$. Peak CPU utilization
increased by $5.87\%$ in relative terms for the larger data set.

In terms of resource utilization, the core algorithms of DeCorus are efficient.
On a data set that consists of close to $150$ million log messages, the core
algorithms each consume in the range of hundreds of MB of memory and around one
to two cores of a modern VM. This is under experimental conditions that reduce
waiting times and increase resource utilization.  

\begin{table}[]
  \caption{Memory consumption in MB of the core algorithms in DeCorus for 280 million and 560 million data points.}
  \begin{tabular}{clll}
                                                    &                           & mean memory             & peak memory            \\ \hline
    \multicolumn{1}{|c|}{\multirow{2}{*}{DC-Z-100\%}} & \multicolumn{1}{l|}{UVAD} & \multicolumn{1}{l|}{446.93} & \multicolumn{1}{l|}{604.22} \\ \cline{2-4} 
    \multicolumn{1}{|c|}{}                            &
    \multicolumn{1}{l|}{MVAD} & \multicolumn{1}{l|}{207.76} & \multicolumn{1}{l|}{418.22} \\ \hline \hline
    \multicolumn{1}{|c|}{\multirow{2}{*}{DC-Z-200\%}} & \multicolumn{1}{l|}{UVAD} & \multicolumn{1}{l|}{477.36} & \multicolumn{1}{l|}{671.73} \\ \cline{2-4} 
    \multicolumn{1}{|c|}{}                            & \multicolumn{1}{l|}{MVAD} & \multicolumn{1}{l|}{240.06} & \multicolumn{1}{l|}{415.65} \\ \hline
  \end{tabular}
  \label{table:performance:memory}
\end{table}

\begin{table}[]
  \caption{CPU utilization in \% of CPU of the algorithms in DeCorus for 280 million and 560 million data points.}
  \begin{tabular}{clll}
                                                    &                           & mean CPU            & peak CPU            \\ \hline
    \multicolumn{1}{|c|}{\multirow{2}{*}{DC-Z-100\%}} & \multicolumn{1}{l|}{UVAD} & \multicolumn{1}{l|}{148.40} & \multicolumn{1}{l|}{614.27} \\ \cline{2-4} 
    \multicolumn{1}{|c|}{}                            &
    \multicolumn{1}{l|}{MVAD} & \multicolumn{1}{l|}{98.23} & \multicolumn{1}{l|}{322.57} \\ \hline \hline
    \multicolumn{1}{|c|}{\multirow{2}{*}{DC-Z-200\%}} & \multicolumn{1}{l|}{UVAD} & \multicolumn{1}{l|}{144.49} & \multicolumn{1}{l|}{661.96} \\ \cline{2-4} 
    \multicolumn{1}{|c|}{}                            & \multicolumn{1}{l|}{MVAD} & \multicolumn{1}{l|}{102.62} & \multicolumn{1}{l|}{341.51} \\ \hline
  \end{tabular}
  \label{table:performance:cpu}
\end{table}

\subsubsection{Runtime}
\label{experiments:performance:runtime}

\begin{table}[]
  \caption{Processing time in seconds of the two core algorithms for a data set with 280 million and 560 million data points over a period of 200 days.}
  \begin{tabular}{|l|c|c|c|}
  \hline
       & DC-Z-100\% & DC-Z-200\% & Increase \\ \hline
  UVAD & 415.18 sec & 768.81 sec & 85.18\%  \\ \hline
  MVAD & 308.49 sec & 581.24 sec & 88.42\%  \\ \hline
  \end{tabular}
  \label{table:performance:time}
\end{table}

As shown in Table~\ref{table:performance:time}, both the UVAD and MVAD algorithm
exhibit similar scalability in their runtime as we increase the volume of data
points. The core processing time of the UVAD algorithm for the smaller data set
consisting of $280$ million data points is $415.18$ seconds. This increases to
$768.81$ seconds for the data set containing twice the number of data points for
the same time period of $200$ days. This represents an increase in the overall
runtime of the UVAD algorithm of $85.18\%$. For the MVAD algorithm, we observe
an increase in core processing time from $308.49$ seconds to $581.24$ seconds,
which represents a relative increase of $88.42\%$. The time spent to analyze
$1,000$ data points is $1.36$ ms for the UVAD algorithm and $23.83$ ms for the
MVAD algorithm on the larger data set. 

The resource requirements of the core algorithms in DeCorus are modest and
increase slowly. Our measurements also confirm that the runtime of the core
algorithms in DeCorus increases in a sub-linear fashion when doubling the number
of data points.

\subsection{Incident Detection Accuracy}
\label{experiments:accuracy}

\subsubsection{Experiment Design}
\label{experiments:accuracy:design}

We examine how well DeCorus can compete with a number of other MVAD approaches
in terms of incident detection accuracy when applied to real-world data.

\begin{table}[]
  \caption{The multivariate anomaly detection techniques used in our comparative
  evaluation.}
  \begin{tabular}{|l|l|l|l|l|}
    \hline
    Approach & Type & Learning & Big-O \\
    \hline
    \hline
    DeCorus & statistical & unsupervised & linear \\
    \hline
    IsolationForest & classification & unsupervised & linear \\
    \hline
    One-Class SVM & classification & unsupervised & cubic \\
    \hline
    LOF & neighbor & unsupervised & quadratic \\
    \hline
    PCA & subspace & unsupervised & cubic \\
    \hline
    Clustering & clustering & semi-supervised & cubic \\
    \hline
  \end{tabular}
  \label{table:accuracy:techniques}
\end{table}

We have selected five techniques (Table~\ref{table:accuracy:techniques}) to
compare DeCorus with, each of which is representative of one of the types of
approaches we have reviewed in Section~\ref{motivation:mvads}.  We focused on
unsupervised learning techniques. However, for the agglomerative hierarchical
clustering algorithm~\cite{https://doi.org/10.2307/2346439}, we used
anomaly-free data for to be able to detect deviations from normal data during
the validation phase.  We did not select another statistical technique to reduce
the already significant implementation effort.

We ran DeCorus as a set of Docker containers implementing the steps of our log
anomaly detection pipeline (cf. Section~\ref{decorus:logad}) on the VM
described in Section~\ref{experiments:performance}.  For the other techniques,
we used the loglizer toolkit~\cite{DBLP:conf/issre/HeZHL16}, which implements a
number of log anomaly detection models. We modified loglizer to work with our
data sets and to integrate some models from implementations in the scikit-learn 
library~\cite{scikit-learn}. We used a server with $196$ GB of RAM to be able to
run the experiments on the loglizer models, which are batch-processing
implementations that expect data to be available in memory.

The data sets we use to evaluate the selected anomaly detection techniques
consist of the raw syslog messages and incident tickets for four cloud service
provider data centers collected over a period of six months
(Table~\ref{table:accuracy:datasetsold}).  The syslog messages are generated by
network devices under production workloads. The four data centers host $4,822$
network devices that have generated more than $1.5$ billion syslog messages in
this time period.  The incident tickets cover a wide range of issues, such as
increased latency on communications links, high error rates on devices, power
loss, network switch reloads, etc.  We removed incidents that were known not to
be visible in syslog messages. Some incidents are about a loss in redundancy
that remains transparent to customers, but NREs prefer to be notified about
these in order to take remedial action. It is to be noted that all incidents 
were acknowledged as real incidents by the NREs.  The number of incidents used
in our experiments is $601$.  

\begin{table}[]
  \caption{The number of devices, syslog messages and confirmed network 
  incidents for four data center networks for a period of six months.}
  \begin{tabular}{|l|c|c|c|c|}
    \hline
      & Devices & Syslogs & \multicolumn{2}{c|}{Incidents} \\
    \hline
      &         &         & Train       & Test \\
    \hline
    DC-A & $75$ & $123,300,724$ & $106$ & $111$ \\
    \hline
    DC-B & $925$ & $263,076,901$ & $59$ & $95$ \\
    \hline
    DC-C & $1,329$ & $349,027,519$ & $75$ & $59$ \\
    \hline
    DC-D & $2,493$ & $857,121,960$ & $45$ & $51$ \\
    \hline
  \end{tabular}
  \label{table:accuracy:datasetsold}
\end{table}

We use three metrics to estimate the accuracy of each anomaly detector:
\emph{precision}, \emph{recall} and \emph{$F_1$-score}. 

\begin{equation}
  Precision = \frac{\text{True Positives}}{\text{True Positives + False Positives}}
\end{equation}

\begin{equation}
  Recall = \frac{\text{True Positives}}{\text{True Positives + False Negatives}}
\end{equation}

\begin{equation}
  F_1 = 2 \cdot \frac{precision \cdot recall}{precision + recall}
\end{equation}

A \emph{True Positive} is a detected anomaly that matches an incident. We count
a match when the anomaly has occurred in the same data center as the incident
and within five minutes of either side of the disruption start time.  We did not
match on the devices affected in an incident as the alternative techniques
mainly identified the anomalous time interval and not all tickets specify the
affected devices. A \emph{False Positive} is a detected anomaly that does not
match an incident (i.e.~a false alert).  And finally, a \emph{False Negative} is
a time interval with at least one incident, for which no anomaly was detected. 

The \emph{precision} of an anomaly detector tells us what proportion of the
anomalies, out of all the anomalies it has detected, do match actual incidents.
The \emph{recall} of an anomaly detector, in turn, quantifies how many of the
actual incidents were detected correctly as anomalies. Finally, the $F_1$-score
is the harmonic mean of these two metrics.  An anomaly detector with high recall
detects most of the positive samples and if it also has high precision, then it
does not flag many samples that should not be detected. The $F_1$-score allows
us to relate these two values in a single number.

We have performed hyperparameter optimization (HPO) for all loglizer models.  We
tried to select relevant hyperparameters with reasonable ranges of values per
model.  We picked $50$ combinations of hyperparameter values from this space at
random and determined the hyperparameter value combination that achieved the
maximum mean $F_1$ score across all data centers, weighted by number of
incidents per data center.  In order to be able to complete the experiments
within a reasonable amount of time, we limited the training time per
hyperparameter value combination to $10$ minutes.  For some of the
higher-complexity models, we increased the training time to four hours.

The accuracy was measured as follows. The input to all models consisted of the
event count vectors that we obtained through pre-processing of the raw syslog
messages (cf. Section~\ref{experiments:performance}). We used a $50-50$
train-test split, in which each anomaly detector is allowed to run on the first
three months of data and is then evaluated on the second three months. This
allows the unsupervised models to adjust any internal parameters before being
evaluated. Clustering was trained on an anomaly-free version of the first three
months of data. We evaluated the output of each model through a scoring function
that applies the definitions discussed earlier. 

\subsubsection{Experiment Results}
\label{experiments:accuracy:results}

\begin{figure}
  \begin{tikzpicture}[scale=0.75]
    \begin{axis}[
        symbolic x coords={DeCorus, LOF, PCA, Isolation Forest, Clustering, OC-SVM},
        xtick=data,
        nodes near coords,
        ylabel={$F_1$ score in \%},
        ybar=1pt,
        x tick label style={rotate=-45},
      ]
        \addplot[ybar,fill=Dark2-D,draw=Dark2-D] coordinates {
            (DeCorus,   6.7)
            (LOF, 2.9)
            (PCA, 2.6)
            (Isolation Forest,  2.2)
            (Clustering,   1.9)
            (OC-SVM, 1.1)
        };
    \end{axis}
  \end{tikzpicture}
  \caption{The weighted mean $F_1$ scores.}
  \label{fig:accuracy}
  \end{figure}
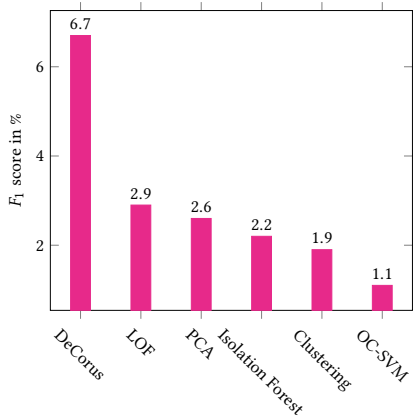

The results of our experiments are summarized in Table~\ref{table:accuracy:f1}
and the weighted mean $F_1$ scores are also compared in
Figure~\ref{fig:accuracy}. The $F_1$ scores per data center are computed as the
harmonic mean of precision and recall. The weighted mean value (WM) is
calculated as the mean of the $F_1$ scores per individual data center, weighted
by the proportion of incidents in each data center. DeCorus achieves the highest
$F_1$ score out of all the anomaly detectors with a WM $F_1$ score of $6.7\%$.
Its accuracy is more than twice that of its closest competitor, LOF, which
achieved an $F_1$ score of $2.9\%$, almost $57\%$ less. With four hours of time
per hyperparameter value combination to process three months of data, LOF
completed $9 - 20$ of the combinations.  LOF was able to detect a greater
proportion of incidents correctly as indicated by its recall of $21\%$ versus a
recall of $7.8\%$ for DeCorus. However, its higher recall seems to come at the
expense of much lower precision: LOF scores $1.6\%$ versus $13.3\%$ for DeCorus.
LOF raises more false positives. LOF achieves its highest $F_1$ score for the
smallest data center (DC-A), on which it outperforms DeCorus, but we observe
decreasing $F_1$ scores for LOF as the size of the data center increases. This
observation holds true for all anomaly detectors apart from DeCorus.

PCA managed to complete processing the first three months of data for all $50$
hyperparameter value combinations when given four hours. Its $F_1$ score of
$2.6\%$ is $61.2\%$ less than that of DeCorus. PCA achieves very high recall
($\> 90\%$) on the two largest data centers, but given its low precision this
does not translate to high $F_1$ scores.  LOF is able to achieve more balanced
('harmonic') accuracy in terms of its precision and recall for the larger data
centers. 

Isolation Forest was able to complete learning on the train data set for a
majority of its hyperparameter settings within the ten-minute limit.  It
achieved an $F_1$ score of $2.2\%$ and was able to detect a greater proportion
of incidents correctly with a recall of $13.4\%$ versus a recall of $7.8\%$
observed for DeCorus.  Again, this seems to come at the expense of much lower
precision (Isolation Forest: $1.3\%$, DeCorus: $13.3\%$).  We observe a spike in
recall for DC-C of $39.7\%$, which we cannot explain. Overall, DeCorus is more
than three times more accurate, in terms of $F_1$ scores.

\begin{table}[]
  \caption{\emph{$F_1$} score, \emph{P}recision and \emph{R}ecall 
  by data center and as
  mean values weighted by the proportion of incidents per data center. The HPO
  column gives the number of hyperparameter configurations for which learning
  was completed. LOF, PCA and OC-SVM were given four hours per
  hyperparameter configuration. No HPO was performed for 
  DeCorus.}
  \label{table:accuracy:f1}
  \begin{tabular}{|l|l|l|l|l|l|l|}
  \hline
                                                                                       &
                                                                                       &
                                                                                       \textbf{F1
                                                                                       \%}
                                                                                       &
                                                                                       \textbf{P \%}
                                                                                       &
                                                                                       \textbf{R \%}
                                                                                       &
                                                                                       \textbf{HPO}
                                                                                       &
                                                                                       \textbf{$\Delta
                                                                                       F_1
                                                                                       \%$}
                                                                                       
                                                                                       \\
                                                                                       \hline
  \multirow{5}{*}{\textbf{DeCorus}}
  & DC-A        & $3.5$       & $21.1$               & $1.9$ &  & 
  \\ \cline{2-7} 
  & DC-B        & $8.2$ & $11.0$ & $6.5$           & &
  \\ \cline{2-7} 
  & DC-C        & $6.8$ & $7.7$ & $6.0$ & &
  \\ \cline{2-7} 
  & DC-D        & $10.8$ & $6.9$ & $25.0$ & &
  \\ \cline{2-7}
                                                                                       &
                                                                                       \textbf{WM}
                                                                                       &
                                                                                       \textbf{$6.7$}
                                                                                       &
                                                                                       \textbf{$13.3$}
                                                                                       &
                                                                                       \textbf{$7.8$}
                                                                                       &
                                                                                       &
                                                                                       \textbf{$0.0$}
                                                                                       \\
                                                                                       \hline
                                                                                       \hline
  \multirow{5}{*}{\textbf{LOF*}}
  & DC-A  &  $5.5$  &  $3.0$  & $33.6$ & $20$ &      \\ \cline{2-7} 
  & DC-B        &  $2.0$           &  $1.1$  & $16.8$ & $11$ &              \\ \cline{2-7} 
  & DC-C        &  $1.1$           &  $0.6$  & $12.1$ & $9$ &              \\ \cline{2-7} 
  & DC-D        &  $0.9$           &  $0.5$  & $12.0$ & $10$ &              \\ \cline{2-7}
                                                                                       &
                                                                                       \textbf{WM}
                                                                                       &
                                                                                       \textbf{$2.9$}
                                                                                       &
                                                                                       \textbf{$1.6$}
                                                                                       &
                                                                                       \textbf{$21.0$}
                                                                                       &
                                                                                       &
                                                                                       \textbf{$-56.7$}
                                                                                       \\
                                                                                       \hline
                                                                                       \hline
  \multirow{5}{*}{\textbf{PCA*}}
  & DC-A        & $5.5$  & $5.0$ & $6.2$ & $50$  &  \\ \cline{2-7} 
  & DC-B        & $1.2$  & $0.6$ & $1.0$ & $50$  &  \\ \cline{2-7} 
  & DC-C        & $0.8$  & $0.4$ & $94.8$  & $50$  &  \\ \cline{2-7} 
  & DC-D        &  $0.7$ & $0.4$  & $92.0$ & $50$  &  \\ \cline{2-7}
                                                                                      &
                                                                                      \textbf{WM}
                                                                                      &
                                                                                      \textbf{$2.6$}
                                                                                      &
                                                                                      \textbf{$2.1$}
                                                                                      &
                                                                                      \textbf{$64.8$}
                                                                                      &
                                                                                      &
                                                                                      \textbf{$-61.2$}
                                                                                      \\
                                                                                      \hline
                                                                                      \hline
  \multirow{5}{*}{\textbf{\begin{tabular}[c]{@{}l@{}}Isolation\\
  Forest\end{tabular}}} & DC-A        & $4.7$ & $2.9$ & $13.3$ &
  50            & \\ \cline{2-7} 
  & DC-B        & $0.9$ & $0.6$              & $2.7$ & 36 &
  \\ \cline{2-7} 
  & DC-C        & $1.1$ & $0.5$ & $39.7$ & 30 &
  \\ \cline{2-7} 
  & DC-D        & $0.5$ & $0.3$ & $3.0$ & 28 &
  \\ \cline{2-7}
                                                                                       &
                                                                                       \textbf{WM}
                                                                                       &
                                                                                       \textbf{$2.2$}
                                                                                       &
                                                                                       \textbf{$1.3$}
                                                                                       &
                                                                                       \textbf{$13.4$}
                                                                                       &
                                                                                       &
                                                                                       \textbf{$-67.2$}
                                                                                       \\
                                                                                       \hline
                                                                                       \hline

  \multirow{5}{*}{\textbf{Clustering}}                                                 & DC-A        & $4.1$       & $3.0$               & $6.7$           & 31 &           \\ \cline{2-7} 
                                                                                       & DC-B        & $1.1$       & $0.6$              & $31.9$           & 13 &           \\ \cline{2-7} 
                                                                                       & DC-C        & $0.8$       & $0.4$              & $9.5$           & 13  &          \\ \cline{2-7} 
                                                                                       & DC-D        & $0.0$           & $0.0$                  & $0.0$               & 16 &           \\ \cline{2-7}
                                                                                       &
                                                                                       \textbf{WM}
                                                                                       &
                                                                                       \textbf{$1.9$}
                                                                                       &
                                                                                       \textbf{$1.3$}
                                                                                       &
                                                                                       \textbf{$13.7$}
                                                                                       &
                                                                                       &
                                                                                       \textbf{$-71.6$}
                                                                                       \\
                                                                                       \hline
                                                                                       \hline

  \multirow{5}{*}{\textbf{OC-SVM*}}                                                    & DC-A        & $1.4$       & $0.7$              & $71.6$           & 47 &           \\ \cline{2-7} 
                                                                                       & DC-B        & $1.3$       & $0.6$              & $97.8$           & 1  &           \\ \cline{2-7} 
                                                                                       & DC-C        & $0.8$       & $0.4$              & $43.1$           & 3  &           \\ \cline{2-7} 
                                                                                       & DC-D        & $0.7$       & $0.4$              & $85.0$            & 1 &            \\ \cline{2-7}
                                                                                       &
                                                                                       \textbf{WM}
                                                                                       &
                                                                                       \textbf{$1.1$}
                                                                                       &
                                                                                       \textbf{$0.6$}
                                                                                       &
                                                                                       \textbf{$76.3$}
                                                                                       &
                                                                                       &
                                                                                       \textbf{$-83.6$}
                                                                                       \\
                                                                                       \hline
  \end{tabular}
  \end{table}

Clustering was able to complete training for a good number of hyperparameter
settings and scores high on recall for DC-B with almost $32\%$, but its
precision is well under $1\%$. Across all four data centers, its $F_1$ score was
$1.9\%$.  It was not able to detect any incidents for the largest data center.
DeCorus achieved an $F_1$ score that is about $3.5$ times higher. 

OC-SVM completed training for almost all hyperparameter value combinations for
the smallest data center, but we only obtained results for up to three settings
on the larger ones. While its recall was among the highest observed at $76.3\%$,
its corresponding precision was only $0.6\%$, which impedes its ability to
generate accurate alerts. Compared to DeCorus, OC-SVM achieves about
$\frac{1}{6}$ of the $F_1$ score. 

We find that DeCorus performs particularly well for the largest data
center, whereas most of the other anomaly detectors achieve their lowest scores
on DC-D. This is an important distinction as the NREs typically need more
support to find relevant incidents in a larger system.  We also observe that the
alternative anomaly detectors tend to achieve higher recall than DeCorus. While
the NREs would ideally want a tool that achieves both very high precision and
recall, we have learnt that they tend to prefer higher precision as one of the
challenges they face is to find the most important alerts in a large set.

\subsubsection{Discussion}
\label{experiments:accuracy:dicussion}

Even DeCorus, which outperforms the other anomaly detectors, displays low
overall accuracy. It detects only one out of every ten incidents and seems to
detect many anomalies where there is no incident ticket. When an anomaly is
detected without a matching incident, we cannot know with certainty that there
was no issue at the time that was simply not recorded in an incident ticket.
This could increase false positives. Similarly, we removed incident tickets we
knew were not detectable via syslog messages, but some such incidents may remain
and increase the false negative rate. Furthermore, we have performed our
evaluation on a single data set. In spite of these concerns, we note that the
syslog messages and incident tickets were from production data centers collected
over a period of six months.  All incidents were confirmed by the NREs to be
real issues as part of their workflow. Each anomaly detector was tested on $300$
incidents.  The use of a large, real-world data sets increases our confidence
that our anomaly detection quality estimates are accurate and that the observed
accuracy is representative of what many unsupervised anomaly detectors can
achieve on large, real-world data sets.

The measures we use as accuracy estimates describe the effect we are interested
in. Precision and recall quantify the ability of an anomaly detector to detect a
high proportion of incidents correctly without raising many needless alerts.  It
can be misleading to consider these metrics in isolation. In our domain, there
are many more benign cases than actual incidents and so an AD that mainly
guesses there to be an anomaly, would identify most real incidents correctly and
achieve high recall, but would probably exhibit low precision. Precision is
important in our domain as too many false alerts may lead NREs to miss true
alerts. Computing the harmonic mean in the form of the $F_1$ score, gives us a
balanced estimate to compare the accuracy of different anomaly detectors.

How can DeCorus, with an $F_1$ score of $6.7\%$, provide benefit in practice?
The NREs reviewed DeCorus alerts over a period of $12$ months before deciding to
deploy it for production monitoring. They concluded that it surfaced relevant
issues without raising too many alerts. They use DeCorus as a kind of
\emph{zero-day} detection tool in addition to the other syslog-based alerting
tools and have been able to add rules for new kinds of incidents to their
deterministic rule base.  DeCorus provides value to the NREs by uncovering
interesting issues in an automated manner. 

We were not able to identify any benefit in terms of accuracy when using
algorithms with super-linear processing cost. 

Our experiments cannot answer if the results would change with more extensive
training time and a larger space of hyperparameter values.  We had to impose
limits on the number of hyperparameter value combinations and the training time
per combination to be able to make experimentation with multiple anomaly
detectors feasible.  However, $50$ combinations of hyperparameter values and up
to four hours of training time per combination are not an insignificant amount
of exploration.

We find that a linear-complexity algorithm based on statistical techniques can
achieve anomaly detection accuracy that is superior to higher-complexity
algorithms and semi-supervised use of clustering. DeCorus achieves the best mix
of precision and recall according to the observed $F_1$ scores and works well
even for the largest data center in our data set. We find that large systems
pose a challenge for all examined anomaly detectors.  While unsupervised
learning is an attractive feature, it is questionable whether a fully
unsupervised approach can achieve high levels of incident detection accuracy in
large systems that produce a high volume of noisy data.

\section{Related Work}
\label{relatedwork}

We are grateful to have been able to use the
Loglizer~\cite{DBLP:conf/issre/HeZHL16} toolkit, which facilitates the
comparison of log anomaly detection approaches. We used a real-world data set
with a greater number of log messages and a smaller proportion of confirmed
incidents than was available in~\cite{DBLP:conf/issre/HeZHL16}. This may
contribute to the different estimate of accuracy we have observed. DeCorus
differs from LogCluster~\cite{lin:mslogcluster:2016} by being
unsupervised and not requiring transaction identifiers in log messages. 
~\cite{xu:2009} is an example of using
PCA to identify issues in logs. The approach requires availability of the source
code of the monitored components, which is not a given in systems that contain
many third-party components.~\cite{10.1145/2783258.2788624} detects anomalies in
individual time series and then combines anomaly scores in a hierarchical
process to find higher-level anomalies. Grouping is not spatio-temporal as in
DeCorus, but is instead based on similarity of signals. In addition, DeCorus
makes use of more domain knowledge (e.g., component criticality weights) in an
attempt to reduce false positives. Finally, in our work we have been able to
contribute the results of a rigorous evaluation based on a real-world data set.

\section{Conclusions and Future Work}
\label{conclusions}

We have described DeCorus, a statistical multivariate anomaly detector that
addresses many of the requirements we have identified as important for use in
large systems.  DeCorus is computationally efficient, operates in an
unsupervised manner, adapts itself to changes in the system, can take temporal
relationships in the data into account and identify contributions of components
to the overall system anomaly score. It makes use of readily-available types of
domain knowledge that can often be provided by NREs to obtain a tree-like
representation of a system and capture information about the criticality of
components and event types.  We described its implementation in an online log
anomaly detection tool used for the analysis of network device syslog messages
at a cloud service provider.

We have characterized the computational requirements of DeCorus in terms of its
resource utilization and runtime and examined its scalability.  Furthermore, we
have compared the incident detection accuracy of DeCorus and five alternative
anomaly detectors.  We used real-world data sets consisting of network device
syslog messages and confirmed incident tickets collected from four cloud data
center networks.

We found that the resource requirements of DeCorus are modest and were able to
confirm sub-linear complexity.  DeCorus achieved accuracy that is three times
higher than that of its closest competitor. However, the real-world data set we
used posed a challenge to all anomaly detectors, including DeCorus.  This may be
representative of the accuracy unsupervised anomaly detection can achieve on
real-world, noisy data.

It is difficult to build highly-accurate failure detection based entirely on
unsupervised multivariate anomaly detection. Accordingly, we want to look into
hybrid approaches that allow us to benefit from some of the advantages of
unsupervised approaches, while making efficient use of small amounts of labeled
data in the form of incident tickets and user-provided feedback to improve
accuracy.

\bibliographystyle{ACM-Reference-Format}
\bibliography{decorus}

\end{document}